\documentclass[11pt]{article}

\usepackage[final]{acl}
\usepackage{times}
\usepackage{latexsym}

\usepackage[T1]{fontenc}
\usepackage[utf8]{inputenc}

\usepackage{microtype}

\usepackage{inconsolata}

\usepackage{graphicx}
\usepackage{amssymb}

\usepackage{ulem}
\usepackage{booktabs}
\usepackage{multirow}
\usepackage{array}
\usepackage{enumitem}
\usepackage{xcolor}
\usepackage{soul}
\usepackage{colortbl}
\usepackage{arydshln}
\usepackage[dvipsnames]{xcolor}
\usepackage{tcolorbox}
\usepackage{amsmath}
\usepackage{amsfonts}
\usepackage{amssymb}
\usepackage{fontawesome5} % 需要 texlive/ miktex 中包含该包

\definecolor{lightgray}{gray}{0.7}
\definecolor{mediumgray}{gray}{0.5}
\definecolor{darkgray}{gray}{0.3}

\title{M$^3$Searcher: Modular Multimodal Information Seeking Agency with Retrieval-Oriented Reasoning}
\author{Xiaohan Yu, Chao Feng, Lang Mei, Chong Chen\textsuperscript{\textcolor{black}{\faEnvelope}} \\
         Huawei Cloud BU, Beijing\\ \texttt{\{yuxiaohan5, fengchao37, meilang1, chenchong55\}@huawei.com}} 

\begin{document}
\maketitle

% \footnotetext[1]{$^{*}$These authors contributed equally to this work.}
% \footnotetext[2]{$^{\dagger}$Corresponding author.}
\footnotetext[2] {\faEnvelope\; Corresponding author.}

\begin{abstract}

Recent advances in DeepResearch-style agents have demonstrated strong capabilities in autonomous information acquisition and synthesize from real-world web environments.
% that large language models (LLMs) can autonomously seek and synthesize information through iterative interaction with real-world web environments.
However, existing approaches remain fundamentally limited to text modality. Extending autonomous information-seeking agents to multimodal settings introduces critical challenges: the specialization-generalization trade-off that emerges when training models for multimodal tool-use at scale, and the severe scarcity of training data capturing complex, multi-step multimodal search trajectories.
% , including the tension between specialized multimodal tool-use and general reasoning capacity, as well as the scarcity of training data that supports long-horizon, multimodal decision-making.
To address these challenges, we propose M$^3$Searcher, a modular multimodal information-seeking agent that explicitly decouples information acquisition from answer derivation. 
M$^3$Searcher is optimized with a retrieval-oriented multi-objective reward that jointly encourages factual accuracy, reasoning soundness, and retrieval fidelity. In addition, we develop MMSearchVQA, a multimodal multi-hop dataset to support retrieval centric RL training. 
Experimental results demonstrate that M$^3$Searcher outperforms existing approaches, exhibits strong transfer adaptability and effective reasoning in complex multimodal tasks.
% \footnote{Code available at \url{https://github.com/yxh-y/M$^3$Searcher}.} 

\end{abstract}

\section{Introduction}
DeepResearch-style agents have recently demonstrated striking proficiency in acquiring and synthesizing information from real-world web environments, as exemplified by OpenAI DeepResearch \cite{openai_deep_research_2024} and Gemini DeepResearch \cite{google_gemini_deep_research_2024}. These advances have spurred a growing research effort to equip large language models (LLMs) with reasoning-intensive search capabilities \cite{shao2025reasonir}. Most approaches leverages reinforcement learning (RL) to train models to interact with web search engines (e.g. Google Search), planning, gathering and synthesizing information through multi-step deliberation  \cite{jin2025search,zheng2025deepresearcherscalingdeepresearch}. 
However, these approaches remain confined to text modality, even though real-world user information needs are inherently multimodal (e.g. visual perception).

Extending autonomous information-seeking agents to multimodal inputs is therefore an essential step for building general intelligent systems. Nevertheless, this transition introduces several fundamental challenges:
(i) \textbf{Specialization-Generalization Trade-off}: 
Training models to internalize multimodal tool-use policies comes at the expense of general reasoning capacity \cite{kalajdzievski2024scaling,li2024revisiting}, yet the effectiveness of multimodal RAG systems critically relies on a backbone model whose core reasoning performance remains robust and uncompromised.
% with uncompromised reasoning capabilities.
% , leading to measurable performance degradation \cite{kalajdzievski2024scaling,li2024revisiting}. 
% This is particularly problematic because high-performing Retrieval-Augmented Generation (RAG) systems require a backbone model with uncompromised reasoning to function effectively.
% However, effective multimodal RAG systems fundamentally depend critically on a strong backbone whose core reasoning capabilities remain uncompromised.
(ii) \textbf{Training Data Scarcity}: 
Existing datasets that capture complex, multi-step search trajectories are primarily designed for evaluation purposes \cite{wei2025browsecomp}, whereas large-scale training corpora provide only shallow reasoning paths, such as InfoSeek \cite{infoseek}. This discrepancy hinders models from developing long-horizon information-seeking strategies.
% in multimodal information seeking.

% Current approaches predominantly pursue this objective by reinforcement learning (RL), such as Search-R1 \cite{jin2025search,zheng2025deepresearcherscalingdeepresearch}.
% The prevailing approaches are driven by reinforcement learning, typified by the Search-R1 paradigm \cite{jin2025search}\st{, recast information seeking from a static, hand-engineered pipeline into an adaptive, closed-loop decision process.}
% This challenge is further compounded by the substantial computational requirements of training: current methods are practically constrained to relatively modest model architectures, such as Qwen-VL-7B. 
% A critical mismatch exists between available datasets and training requirements. 

\begin{figure*}
    \centering
    \includegraphics[width=\linewidth]{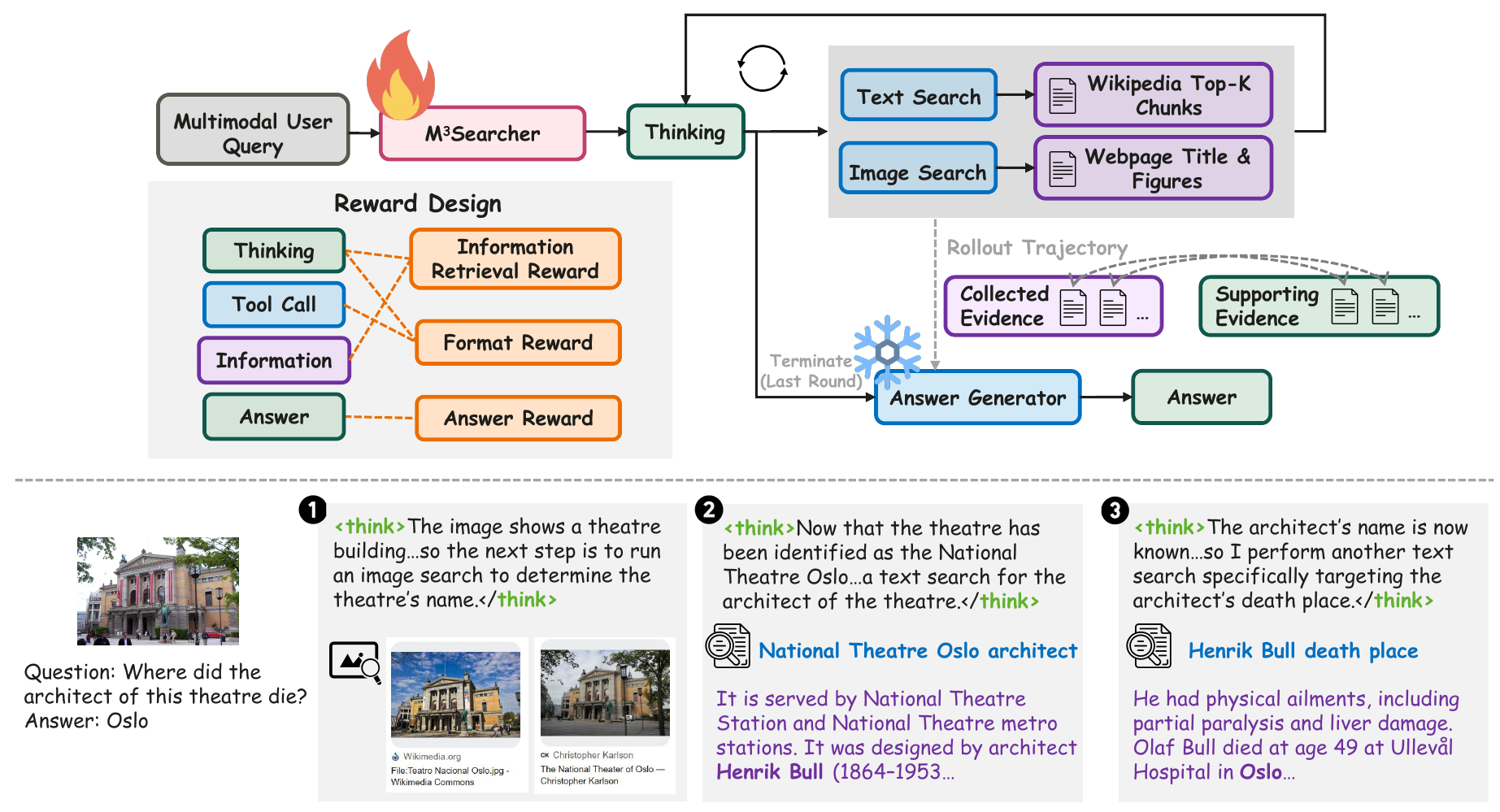}
    \caption{The architecture of M$^3$Searcher.}
    \label{model}
\end{figure*}

To resolve these challenge, and inspired by the modular design of \citet{jiang2025s3}, we decouple the information-seeking process from answer derivation. Specifically, we introduce a lightweight and trainable MLLM, termed \textbf{M$^3$Searcher}, that serves as a dedicated modular multimodal information seeking agency.
Its role is to execute a mulitmodal reasoning-intensive information seeking process \cite{shao2025reasonir}. 
Specifically, it interprets non-textual inputs (e.g. visual recognition, OCR) and dynamically coordinating search strategies across heterogeneous modalities to assemble comprehensive and contextually relevant evidence.
The gathered information is subsequently provided to a downstream answer generator, which performs reasoning over the curated evidence and formulates the final response to the user query.
To effectively train M$^3$Searcher, we further propose a decoupled reinforcement learning framework with the following contributions:
% To endow M$^3$Searcher with robust multimodal reasoning and autonomous information-seeking capabilities, we introduce a decoupled RL framework that achieves the following contributions:
\begin{itemize}[leftmargin=*]
    \item[1.] \textbf{Dataset Construction:} 
    We introduce \textbf{MMSearchVQA}, a dataset demanding rigorous multimodal information seeking. Each instance enforces answer uniqueness and is accompanied by automatically extracted supporting evidence.
    By encompassing a broad spectrum of domains, difficulty levels and search intensities, the dataset encourages the model to learn the distinct control policies required for determining \textit{when} to search, \textit{what} to query, and \textit{how} to integrate external knowledge.
    \item[2.] \textbf{Decoupled Multimodal Information Seeking:} M$^3$Searcher focuses exclusively on optimizing heterogeneous search scheduling for maximizing information acquisition. To realize this, we introduce a specialized "expert answer generator" tool, which is triggered only once the context is deemed sufficient and well-grounded for the following reasoning process.
    This modularity allows the search strategies to remain highly adaptive while maintaining the reasoning capacity of a robust backbone within the MRAG system.
    It also renders the generator modality-agnostic, accommodating both pure textual LRMs (e.g., DeepSeek-R1) and MLLMs (e.g., GPT-4o).
    \item[3.] \textbf{Retrieval-Oriented Multi-Objective Reward:} We employ a multi-objective reward modeling framework that jointly optimizes answer accuracy, reasoning validity, and retrieval quality. To ensure that the model genuinely grounds its inferences in retrieved evidence rather than exploiting spurious shortcuts, we incorporate a retrieval reward that evaluates the completeness of textual information gathering and the accuracy, and interpretive soundness of visual reasoning.
\end{itemize}

We conduct comprehensive evaluations of M$^3$Searcher across real-world benchmarks to assess its effectiveness. M$^3$Searcher outperforms both prompt-engineered agents and end-to-end trained counterparts. Moreover, it exhibits strong robustness and adaptability, as evidenced by stable performance under multiple transfer scenarios involving variations in search engines and answer generators.

\section{M$^3$Searcher}

\subsection{Task Formulation}
We consider a multimodal query $(v, q)$ where $v$ is the visual component and $q$ is the textual component, with its ground-truth answer $a$. A trainable MLLM is formalized as an information-seeking agent $\mathcal{F}$, which engages in iterative interaction with a multimodal tool set $\mathcal{T}$. Through a sequence of tool invocations and intermediate reasoning steps, the agent incrementally acquires task-relevant evidence and integrates the retrieved information to derive the final answer:
\begin{align}
\mathcal{F}(q, v, \mathcal{T}) \rightarrow a.
\end{align}

% To operationalize this process, M$^3$Searcher processes the query through three core states. In the \textcolor{RoyalBlue}{\textbf{\texttt{Think}}} state, the model performs fine-grained inspection of the visual input $v$ and conducts cross-modal reasoning by integrating visual and textual cues into a coherent situational understanding. When the current internal knowledge is insufficient, the agent transitions to the \textcolor{YellowOrange}{\textbf{\texttt{Tool\_Call}}} state, in which it dynamically invokes external multimodal tools to retrieve supplementary evidence from real-world sources. The returned results are encapsulated as \textcolor{BrickRed}{\textbf{\texttt{Information}}} and fed back into the reasoning loop to refine subsequent decisions.

% M$^3$Searcher processes the query through three core operational states. In the \textcolor{RoyalBlue}{\textbf{\texttt{Think}}} state, the model conducts a fine-grained inspection of the visual component $v$ and performs contextual inference across modalities, integrating visual and textual cues to construct a coherent situational understanding. When additional information is required, M$^3$Searcher transitions to the \textcolor{YellowOrange}{\textbf{\texttt{Tool\_Call}}} state, dynamically invoking external tools to retrieve supplementary evidence from real-world sources. The retrieved outputs are then encapsulated as \textcolor{BrickRed}{\textbf{\texttt{Information}}}, which re-enters the reasoning loop to refine and expand the model’s understanding

\subsection{Decoupled Agentic MRAG}
Existing MRAG approaches either leverage a large-scale MLLM with elaborate prompt engineering, or train a smaller MLLM end-to-end \cite{geng2025webwatcher,wu2025mmsearch}. This dichotomy introduces a fundamental dilemma. Large-scale MLLMs (e.g. GPT-4o) excel in emergent reasoning, but lack optimization for real-world web integration. Conversely, smaller models (e.g., Qwen2.5-VL-7B) optimized for web search exhibit a concurrent degradation in general reasoning, limiting their utility as a primary backbone for the overall system.
To address this dilemma, and drawing inspiration from modular architectures \cite{jiang2025s3}, we introduce a decoupled MRAG architecture that separates the information-seeking process from the answer generation. 
As depicted in Figure \ref{pipeline}, M$^3$Searcher is solely responsible for comprehending multimodal queries, formulating iterative search strategies across heterogeneous search tools, and determining the optimal search termination point. The final evidential data is then passed to a dedicated answer generation for synthesis.
This architectural decoupling offers two key advantages:
it preserves the reasoning fidelity of the large-scale backbone while allowing for targeted optimization of information-seeking capabilities, and simultaneously removes modality constraints, enabling the use of modality-agnostic generators (e.g., GPT-4o, DeepSeek-R1).
% . First, it preserves the reasoning strength of large-scale backbones while allowing for targeted optimization of multimodal information-seeking capabilities. Second, it removes modality constraints: as long as M$^3$Searcher satisfies the modality requirements of the query, the answer generator can remain modality-agnostic (e.g. GPT-4o, DeepSeek-R1).
% , seamlessly integrating both closed-source models (e.g., GPT-4o) and open-source counterparts (e.g., DeepSeek-R1).

\subsection{Multimodal Tools Implementation}
\label{tools}
 % The search tools interface with web engines using modality-specific queries (e.g., textual or visual) and return structured outputs. The answer generator consumes the trajectory of information-seeking and synthesizes a final response. The iterative retrieval process proceeds until sufficient evidence has been gathered and invokes the answer generator to conclude the task.

We equip M$^3$Searcher with three essential tools: an image search tool, a text search tool, and an answer generator tool. To enable effective RL, we developed a stable and high-concurrency tool environment.
For \texttt{image search}, we integrate the Serper API\footnote{https://serpapi.com/} to perform reverse image retrieval. Given an input image, the API returns visually similar images, together with their corresponding website titles and URLs. Since the Serper API produces highly stable results,  we incorporate a caching mechanism to reduce resource consumption and accelerate the search process.
For \texttt{text search}, we utilize the 2025 wikipedia dump\footnote{https://dumps.wikimedia.org/} as knowledge source. A retrieval–reranking pipeline, built upon the E5 models \cite{wang2022text}, is used to retrieve semantically relevant document chunks given a user query.
Finally, the \texttt{answer expert} tool employs a high-capacity LRM which consumes the trajectory of information-seeking and synthesizes a final response.
% synthesizes information from both image and text retrieval and generates final responses.
% and build a retrieval - reranking pipeline leveraging the E5 models \cite{chen2024bge}. Given a user query, the pipeline returns the semantically relevant document chunks.

\subsection{Decoupled Multi-turn Rollout}
M$^3$Searcher processes the query through three core operational states. In the \textcolor{RoyalBlue}{\textbf{\texttt{Think}}} state, the model conducts a fine-grained inspection of the visual component $v$ and performs contextual inference across modalities, integrating visual and textual cues to construct a coherent situational understanding. When additional information is required, M$^3$Searcher transitions to the \textcolor{YellowOrange}{\textbf{\texttt{Tool\_Call}}} state, dynamically invoking external tools to retrieve supplementary evidence from real-world sources. The retrieved outputs are then encapsulated as \textcolor{BrickRed}{\textbf{\texttt{Information}}}, which re-enters the reasoning loop to refine and expand the model’s understanding
% It performs a fine-grained inspection of the visual component $v$, and engages in contextual analysis and multimodal reasoning in \textcolor{RoyalBlue}{\textbf{\texttt{Think}}} state. When additional evidence is required, the planner dynamically invokes external multimodal tools, enabling direct interaction with the real-world web environment through \textcolor{YellowOrange}{\textbf{\texttt{Tool\_Call}}}. The invoked tools returns responses by  \textcolor{BrickRed}{\textbf{\texttt{Information}}}.
M$^3$Searcher operates iteratively upon these three states, allowing for progressive refinement of its understanding and retrieval strategy. Formally, at each time step $t$, the M$^3$Searcher execution can be represented as a tuple $(\alpha_t, C_t, I_t)$, where $\alpha_t$ represents the reasoning process, $C_t$ is the tool invocation, and $I_t$ is the tool response. The full rollout trajectory can thus be expressed as:
\begin{align}
    \mathcal{T} = \{O_1, \alpha_1, C_1, I_1, \ldots, O_t, \alpha_t, C_t, I_t\}.
\end{align}
Under the decoupled agentic design, the final tool invocation of M$^3$Searcher is required to invoke the answer expert. Consequently, the terminal tool response $I_t$ provides the final answer to the user query $q$:
\begin{align}
    I_t = \mathcal{F}(q).
\end{align}
The prompt governing the rollout procedure is detailed in Appendix \ref{rollout}.
% handle it in a agentic manner which can be formulated with state transition. 

% \textcolor{RoyalBlue}{\textbf{\texttt{Observe}}}: MMPlanner first performs a fine-grained inspection of the visual component $v$, extracting high-confidence semantic information.
%     This involves interpreting the visual content, recognizing relevant elements grounded in its knowledge base, and applying optical character recognition (OCR) when necessary (e.g., when screenshots contain equations or paragraphs of text).
    % \item[2.] \textcolor{YellowGreen}{\textbf{\texttt{Think}}}: MMPlanner engages in contextual analysis and multimodal reasoning, integrating both visual and textual cues to construct a coherent internal representation of the current progress in task solving.
    % \item[3.] \textcolor{YellowOrange}{\textbf{\texttt{Tool\_Call}}}: When additional evidence is required, the planner dynamically invokes external multimodal tools, enabling direct interaction with the real-world web environment and other specialized resources.
    % \item[4.] \textcolor{BrickRed}{\textbf{\texttt{Information}}}: This state captures the responses returned by the invoked tools.

\subsection{Multi-Objective Rewrad Modeling}
The goal of M$^3$Searcher is to perform a multimodal, reasoning-intensive information-seeking process. It must progressively and comprehensively gather relevant evidence across multiple hops to support the downstream generator in generating accurate answers. To achieve this, we formulate a multi-objective, retrieval-oriented RL reward function that jointly optimize accuracy, completeness and relevance of the information acquisition process.

\noindent \paragraph{Format Reward} 
The format reward $R_{format}$ enforces strict compliance with the syntactic and structural constraints specified in the prompt. For example, tool invocations are required to follow the correctly structured parsing format with valid parameterization; and the trajectory must terminate with a call to the answer generator tool. Any deviation from these requirements incurs a strong penalty of an absolute reward of -1.
% the \textcolor{RoyalBlue}{\textbf{\texttt{Think}}} state must be enclosed within the designated \textcolor{RoyalBlue}{<observe>} and \textcolor{RoyalBlue}{</observe>} tags; 
% \begin{align}
% R_{format}(\mathcal{T}) =
% \begin{cases}
% 0, & \text{if the trajectory format is valid}, \\
% -2, & \text{otherwise}.
% \end{cases}
% \end{align}

\paragraph{Answer Reward} 
The answer reward, $R_{answer}$, measures the semantic correctness of the final output $I_t$ with respect to the reference solution. Rather than relying on brittle exact string matching, we employ an LLM-as-Judge evaluation strategy, which confers both flexibility and robustness in cases where multiple equivalent phrasings or semantically consistent answers are acceptable. The complete scoring prompt used for this evaluation is provided in the Appendix.

\begin{figure*}
    \centering
    \includegraphics[width=\linewidth]{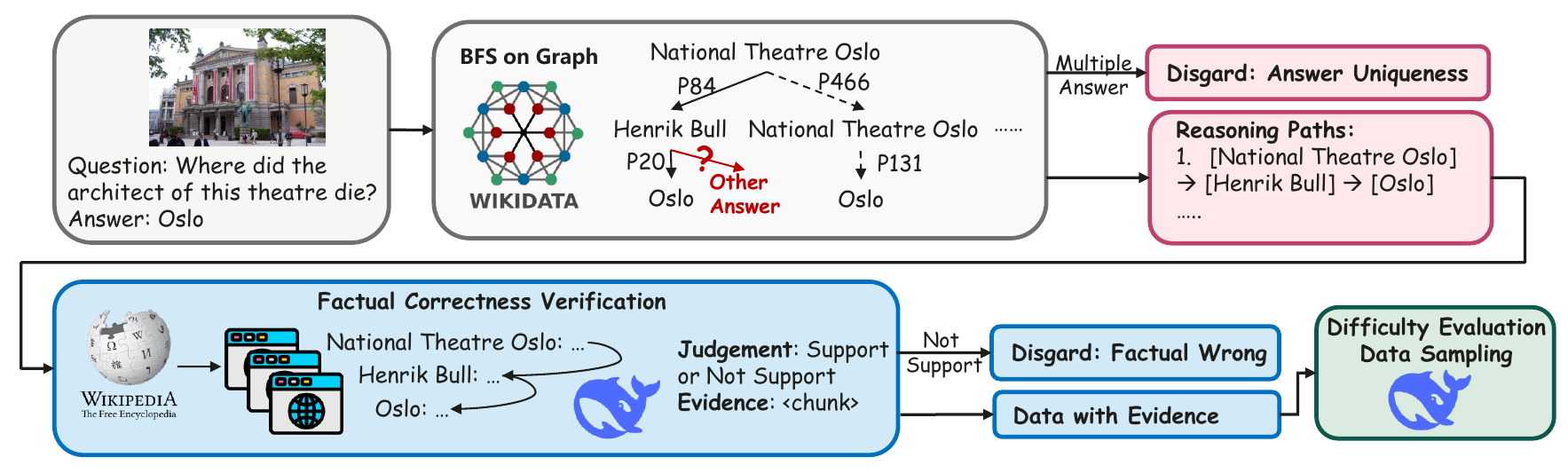}
    \caption{The MMSearchVQA data construction pipeline.}
    \label{pipeline}
\end{figure*}

\begin{figure}
    \centering
    \includegraphics[width=0.50\linewidth]{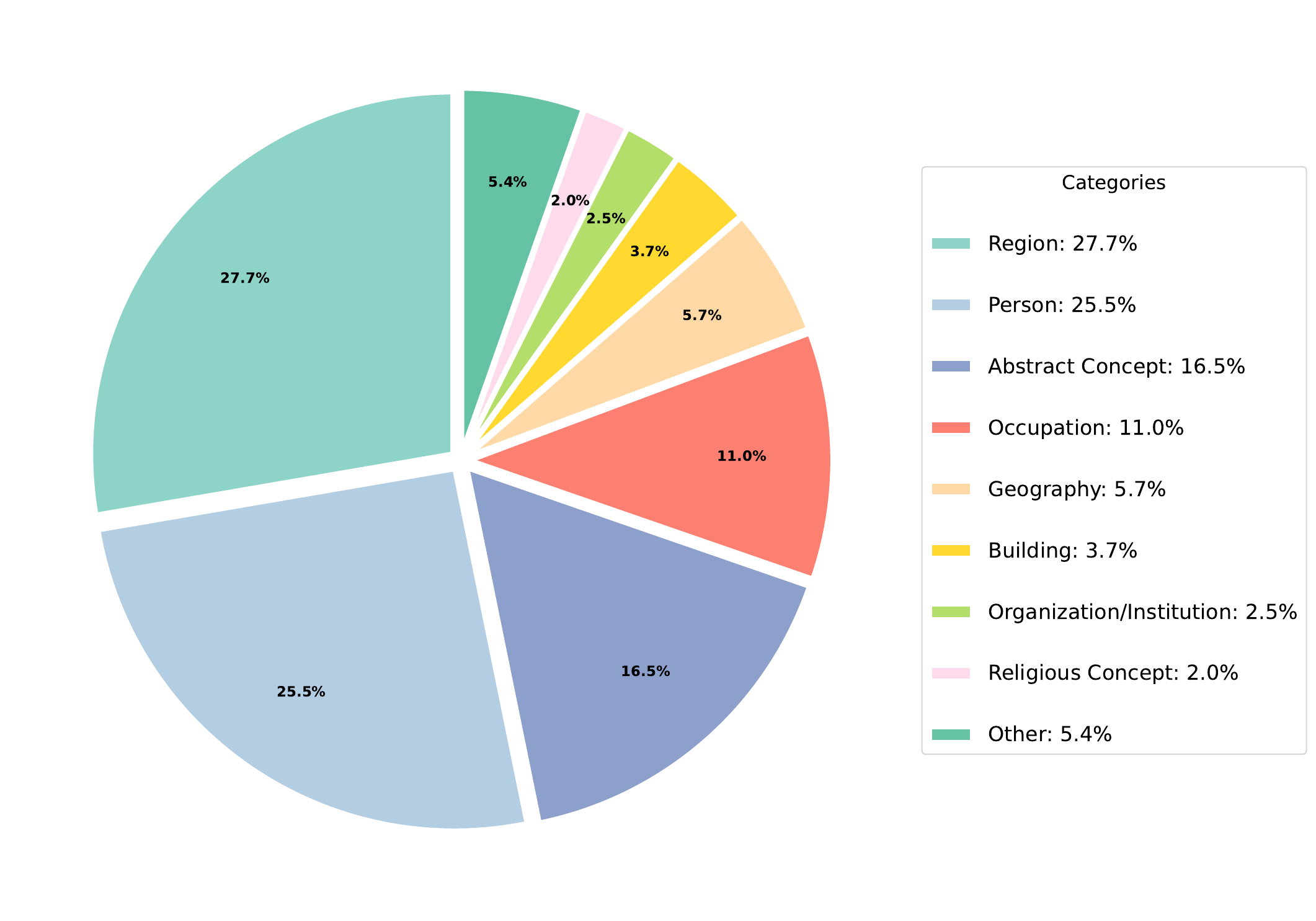}
    \includegraphics[width=0.45\linewidth]{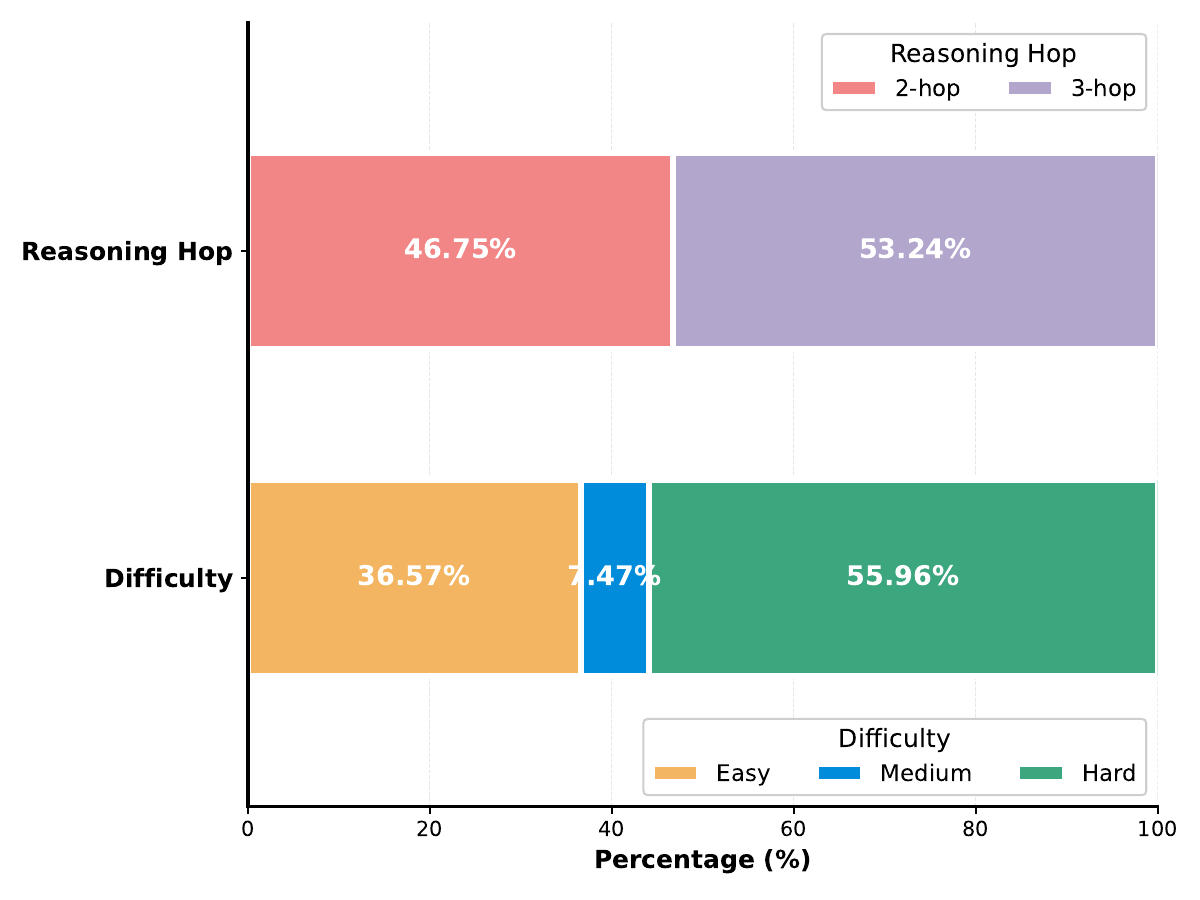}
    \caption{Overview of MMSearchVQA dataset statistics. The left figure summarizes the domain distribution, and the right figure reports the distribution of question difficulty levels and reasoning hops.}
    \label{stati}
\end{figure}

\paragraph{Information Retrieval Reward}
The information retrieval reward is designed to assess the fidelity and completeness of the information acquired to solve a multi-hop user query, independent of the capabilities of the downstream answer generator. The evaluation of this information acquisition is divided by modality.

For the visual modality, when processing retrieved images or relying on internal knowledge within the model’s training cutoff, it may exhibit three distinct behaviors during the \textcolor{RoyalBlue}{\textbf{\texttt{Think}}} state: (1) correctly identifying the key visual elements, (2) demonstrating uncertainty and refraining from explicit recognition while offering a descriptive interpretation, or (3) producing an incorrect recognition. To shape this behavior, we assign graded rewards $R_{ImgRetrieval}$ of 0.5, 0.25, and 0, respectively. This reward structure encourages the model to adopt a more cautious and self-aware strategy when reasoning over visual inputs.
% the planner has two possible actions: it may either attempt direct recognition using its internal knowledge cutoff or provide a descriptive account when uncertain. Both choices are acceptable in principle; however, incorrect recognition is explicitly penalized with a negative reward of $-1$. This penalty encourages the planner to adopt a more cautious strategy when interpreting visual inputs.
For the textual modality, we assess the degree to which the information conveyed to the answer generator aligns with the reference evidence in MMSearchVQA. This metric quantifies whether M$^3$Searcher identifies all necessary pieces of information required for solving the query, thereby enhancing information completeness and mitigating reasoning shortcuts that may yield correct answers without genuine verification (e.g., a builder’s place of death is not always the same as the building’s location).
The textual retrieval reward, denoted as $R_{TextRetrieval}$, is defined as a percentage score ranging from 0 to 0.5, representing the proportion of reasoning hops successfully supported by retrieved evidence. Specifically, we compare each reference evidence against both the \textcolor{BrickRed}{\textbf{\texttt{Information}}} and \textcolor{RoyalBlue}{\textbf{\texttt{Think}}} states.
% For the textual modality, we measure the textual information convey to the answer generator against the golden evidence sources, This metric reflects whether M$^3$Searcher successfully identifies all necessary pieces of information required for solving the query. it improve the information completeness and try to avoid reasoning shortcuts that may just guess right answer without verification (e.g. a building constructor died place does not necessarily equal to the place of this building). $R_{TextRetrieval}$ is a percentage score depending how much of the reasoning hop is successfully supported ranging from 0 to 1, we will compare each golden evidence with both \textcolor{BrickRed}{\textbf{\texttt{Information}}} and the \textcolor{RoyalBlue}{\textbf{\texttt{Think}}} state (for the general simple reasoning can be finished by model common knwoledge).
To ensure robust evaluation, we employ a LLM-as-Judger method to assess both modality reward score:
\begin{align}
    &R_{ImgRetrieval} = LLM(\alpha_i), \\
    &R_{TextRetrieval} = LLM(\alpha_i, C_i).
\end{align}
The detailed judging prompt is provided in the Appendix.
The final reward is:
\begin{align}
    &R = R_{format} + R_{answer} + R_{Retrieve}, \\
    &R_{Retrieve} = R_{TextRetrieval} + R_{ImgRetrieval}.
\end{align}

\subsection{RL Training}
To enhance the model’s capability for information seeking and web-environment interaction within the MRAG framework, we adopt Group-Relative Policy Optimization (GRPO) \cite{grpo}. For each input multimodal question $q$, the current policy $\pi_{\theta}$ samples a group of trajectories ${y_1,\ldots,y_G}\}$. Then the optimization objective of GRPO is formulated as:
% Unlike conventional reinforcement learning strategies that rely on a separate value function, GRPO normalizes rewards within each sampled group, allowing more stable and sample-efficient optimization. 

\begin{equation}
\begin{aligned}
    \mathcal{J}(\theta) &= \mathbb{E}_{i,t}
    \min \biggl[ \rho^{i}_{t} A^{i}_{t}, 
    \operatorname{clip}(
     \rho^{i}_{t},\ 1-\epsilon,\ 1+\epsilon)A^{i}_{t}\biggr] \\
    & - \beta \, \mathbb{D}_{\mathrm{KL}}[\pi_{\theta} \vert \vert \pi_{\text{ref}}],
\end{aligned}
\end{equation}
where $\rho^{i}_{t}$ represents the importance sampling ratio between the updated and previous policies and $A^{i}_{t}$ is an estimator of the advantage at time step $t$:
\begin{align}
    % \rho^{i}_{t}=\frac{\pi_{\theta}(y_{i,t}\mid context)}{\pi_{\text{old}}(y_{i,t}\mid context)} \\
    \quad A^{i}_{t} = \frac{R_i - mean(\{R_i\})}{std(\{R_i\})}.
\end{align}
The hyperparameter $\beta$ controls the KL divergence penalty, constraining the deviation from the reference policy to ensure stable updates. The context for policies includes both model-generated outputs and tool responses. To prevent external knowledge sources from biasing policy learning, we apply a loss mask over all tool-response tokens. This ensures that policy gradients are computed exclusively for LLM-generated tokens, enabling precise optimization of search planning and multimodal information-seeking capabilities within the MRAG system.

\section{MMSearchVQA Dataset}
Existing Visual Question Answering (VQA) datasets typically fall into two categories. Automatically constructed datasets \cite{infoseek,cheng2025simplevqa,wu2025mmsearch,fu2025livevqa}, such as InfoSeek, involve shallow reasoning chains — often limited to two-hop queries solvable through a simple sequence of image search followed by text search. In contrast, manually curated datasets such as MM-BrowseComp \cite{li2025mm} feature more complex, multi-step reasoning but are expensive and difficult to scale. 
To address this limitation, we introduce MMSearchVQA, a dataset designed to foster the development of models for advanced information-seeking reasoning. MMSearchVQA not only requires deeper search and reasoning but also provides explicit supporting evidence that underpin the reasoning and answering processes.

As illustrated in Figure \ref{pipeline}, our dataset is constructed upon ReasonVQA \cite{tran2025reasonvqa}, which is derived from the Wikidata. We first perform a BFS traversal on the Wikidata graph, identifying all potential reasoning chains associated with each question. During this traversal, we discard questions that yield multiple valid answers to ensure answer uniqueness, and we retain only those samples that require at least two reasoning hops.
Following the extraction of candidate reasoning paths, we conduct cross-validation against Wikipedia using the DeepSeek models. Each reasoning hop, including the final answering, must be consistently supported by evidence drawn from relevant Wikipedia content to be both factually accurate and temporally valid. Instances that fail to meet these criteria are excluded. During this verification process, we also extract fine-grained supporting evidence from the corresponding Wikipedia passages for each reasoning step, thus enhancing the interpretability and traceability of the reasoning process.
To further characterize the cognitive difficulty of the resulting dataset, we employ DeepSeek-V3 \cite{guo2025deepseek} to answer each question three times and categorize questions into three levels: easy (all correct), medium (partially correct), and hard (all incorrect). This procedure yields a principled estimate of reasoning complexity across samples.
% , enabling a more systematic assessment of model performance under varying reasoning demands. 
To cultivate an information seeker capable of performing deep and precise searches, we prioritize training data that exhibit deeper information needs and greater reasoning complexity. Accordingly, we downsample easy questions to half the number of hard examples, ensuring a balanced yet challenging dataset. In total, the curated dataset contains 6,000 questions, with comprehensive statistics presented in Figure \ref{stati}.
% approximately 3,000 two-hop and 3,000 three-hop reasoning questions.

\begin{table*}[t]
  \centering
  \caption{The overall performance of M$^3$Searcher compared with baseline approaches. MMSearchVQA is in-domain benchmark with wikipedia based search and other benchmarks are transferred to Google Search Engine. The best performance is highlighted in \textbf{bold}, and the second-best performance is \underline{underlined}.}
  \setlength{\tabcolsep}{8pt}
  \small
\begin{tabular}{llcccc}
    \toprule[1.3pt]
     \multirow{2}{*}{\textbf{Method}} & \multirow{2}{*}{\textbf{Backbone}} & \textbf{In-Domian (Wiki Search)} & \multicolumn{3}{c}{$\rightarrow$ \textbf{Out-Domain (Google Search)}}  \\ \cmidrule(lr){3-3} \cmidrule(lr){4-6}
     & & \textbf{MMSearchVQA} & \textbf{InfoSeek} & \textbf{MMSearch} & \textbf{MRAG-Bench}   \\  
    \midrule 
    \rowcolor{gray!15} \multicolumn{6}{c}{\textbf{No Agency}} \\
    \midrule 
    \multirow{3}{*}{Direct} & Qwen3-VL-235B-A22B & 40.42 & 40.16 & 28.23 & 9.22  \\
     & Qwen2.5-VL-72B  & 43.12 & 35.22 & 15.29 &  14.12  \\
     & Qwen2.5-VL-7B & 31.12 & 23.50 & 11.69 & 8.57 \\
     \arrayrulecolor{gray!30}  % 30% gray
    \midrule
    \arrayrulecolor{black}
    \multirow{3}{*}{RAG} & Qwen3-VL-235B-A22B & \textcolor{black}{29.79}  & 31.75 & 30.83 & \textbf{33.67} \\
     &  Qwen2.5-VL-72B   &  40.37 & 33.26 & \underline{46.15} & 20.08 \\
     &  Qwen2.5-VL-7B & 30.00 & 31.75 & 38.09 & 15.68 \\
    \midrule
    \rowcolor{gray!15} \multicolumn{6}{c}{\textbf{Prompt Engineered Agents}} \\
    \midrule
    \multirow{2}{*}{OmniSearch}  & Qwen2.5-VL-72B  & 45.65 & 40.60 & 15.00 & 27.07  \\
    & Qwen2.5-VL-7B & 22.91 & 25.17 & 22.22 & 23.96 \\
     \arrayrulecolor{gray!30}  % 30% gray
    \midrule
    \arrayrulecolor{black}
    \multirow{2}{*}{CogPlanner}  & Qwen2.5-VL-72B  & \underline{48.37} & \underline{41.72} & 39.77 & 29.12  \\
    & Qwen2.5-VL-7B  & 22.12 & 26.22 & 27.48 & 28.23  \\
    \midrule
    \rowcolor{gray!15} \multicolumn{6}{c}{\textbf{End-to-End Agents}} \\
    \midrule
    \multicolumn{2}{l}{MMSearch-R1-7B$_{Retrain}$} & 31.63 & 20.20 & 7.02 & 27.60 \\
    \multicolumn{2}{l}{MMSearch-R1-7B$_{Release}$} & 20.50 & 37.06 & 12.28 & 19.20  \\
    %  \arrayrulecolor{gray!30}  % 30% gray
    % \midrule
    % \arrayrulecolor{black}
    % \multicolumn{2}{l}{WebWatcher-7B} \\
    \midrule
    \rowcolor{gray!15} \multicolumn{6}{c}{\textbf{Decoupled Agents w/o Training}} \\
    \multicolumn{2}{l}{Qwen3-30B-A3B} &  31.50 & 31.20 & 36.69 & 24.20  \\
    \arrayrulecolor{gray!30}  % 30% gray
    \hdashline[2pt/1pt]
    \arrayrulecolor{black}
    \multicolumn{2}{l}{Qwen2.5-VL-7B} & 34.12 & 33.80 & 36.09 & 27.20 \\
    \midrule
    \rowcolor{gray!15} \multicolumn{6}{c}{\textbf{M$^3$Searcher}} \\
    \midrule
    \multicolumn{2}{l}{Qwen3-30B-A3B } & 54.75 & 39.61 & 55.62 & 24.91  \\
    % \multicolumn{2}{l}{\textbf{MMPlanner-7B} \& Qwen3-32B Generator} & 54.75 & 39.61 & 55.62 & 24.91  \\
    \arrayrulecolor{gray!30}  % 30% gray
    \hdashline[2pt/1pt]
    \arrayrulecolor{black}
    \multicolumn{2}{l}{\textcolor{gray}{$\rightarrow$ \textit{Transfer LRM Answer Generator}}} &  \\
    \multicolumn{2}{l}{\quad DeepSeek-V3} & 56.87 & 40.33 & 60.95 & 29.12  \\
    \multicolumn{2}{l}{\quad DeepSeek-R1} & 59.25 & \textbf{42.50} & \textbf{63.30} & \underline{30.00} \\
    \arrayrulecolor{gray!30}  % 30% gray
    \hdashline[2pt/1pt]
    \arrayrulecolor{black}
    \multicolumn{2}{l}{\textcolor{gray}{$\rightarrow$ \textit{Transfer MLLM Answer Generator}}} &  \\
    \multicolumn{2}{l}{\quad Qwen2.5-VL-7B} & 57.00 & 39.44 & 61.54 & 19.95 \\
    \multicolumn{2}{l}{\quad Qwen2.5-VL-72B} & \textbf{59.50} & 40.20 & 59.17 & 27.20  \\
    \bottomrule[1.3pt]
  \end{tabular}
\label{main}
\end{table*}

\section{Experiments}

\subsection{Experimental Setup}
\paragraph{Datasets} 
We adopt MMSearchVQA dataset as the training corpus. 
% To comprehensively evaluate the generalization capability of MMPlanner, we conduct experiments on both in-domain and out-of-domain benchmarks. 
% The in-domain benchmark is the FVQA test set, while the out-of-domain benchmarks consist of three publicly available VQA datasets: MMSearch \cite{jiang2024mmsearch}, Infoseek \cite{infoseek} and MRAG-Bench \cite{hu2024mrag}.
We evaluate performance on both in-domain and out-of-domain benchmarks. The in-domain evaluation uses the MMSearchVQA test set, while out-of-domain evaluation is conducted on three publicly available VQA datasets: MMSearch \cite{jiang2024mmsearch}, Infoseek \cite{infoseek}, and MRAG-Bench \cite{hu2024mrag}.

\paragraph{Baselines and Metrics}
We compare M$^3$Searcher against four categories of methodologies: (1) No-agency: We directly prompt MLLMs and use a fixed RAG pipeline comprising image retrieval, query rewriting, text retrieval, and answer generation. (2) Prompt-engineered agents: We select OmniSearch \cite{li2024benchmarkingmultimodalretrievalaugmented} and CogPlanner \cite{yu2025unveiling}, both of which coordinate multiple agents via hand-crafted prompts for multimodal reasoning and retrieval. (3) End-to-end agents: We include MMSearch-R1 \cite{wu2025mmsearch} as a representative method optimized through end-to-end RL training. (4) Decoupled agents without specialized training: We employ a decoupled architecture in which Qwen2.5-VL-7B is used for information-seeking operations, without any tuning.
We adopt LLM-as-Judge as the evaluation metric, which is well-aligned with the answer accuracy reward.

\paragraph{Transfer Experiment Settings} For M$^3$Searcher, we conduct two sets of transfer experiments: (1) Search engine transfer: For the MMSearchVQA benchmark, which is built on Wikipedia-based content, we employ our in-house text search tool (described in Section~\ref{tools}) to retrieve the top 10 most relevant text chunks. For other benchmarks based on open-domain web data, we switch to Google Search via the Serper API\footnote{\url{https://serper.dev/}}
, also keeping the top 10 retrieved results. (2) Answer generator transfer: We explore the transfer of answer generators by incorporating models from both the DeepSeek series \cite{guo2025deepseek, liu2024deepseek} and the Qwen-VL series \cite{bai2025qwen25}.

\paragraph{Implementation Details}
For the baseline methodologies, we adopt Qwen3-VL-30B-A3B, Qwen2.5-VL-72B, and Qwen2.5-VL-7B as backbone models. Specifically, MMSearch-R1 employs Qwen2.5-VL-7B for both end-to-end training and inference.
For M$^3$Searcher, we utilize Qwen2.5-VL-7B as the trainable planner and Qwen3-30B-A3B \cite{yang2025qwen3} as the answer generator during training. 
% The wiki search segments the wikipedia webpage content into chunks of 2,000 tokens with a 200-token overlap and retrieves the 10 most relevant chunks, while the image search returns one image along with the top 30 related webpage titles.
% During training, the maximum reasoning and tool calling iteration is capped at 10. The GRPO  group size is set to 8 to balance exploration and convergence.
Verl \cite{sheng2024hybridflow} is used for multi-turn RL training. 

\begin{figure}[t]
    \centering
    \includegraphics[width=\linewidth]{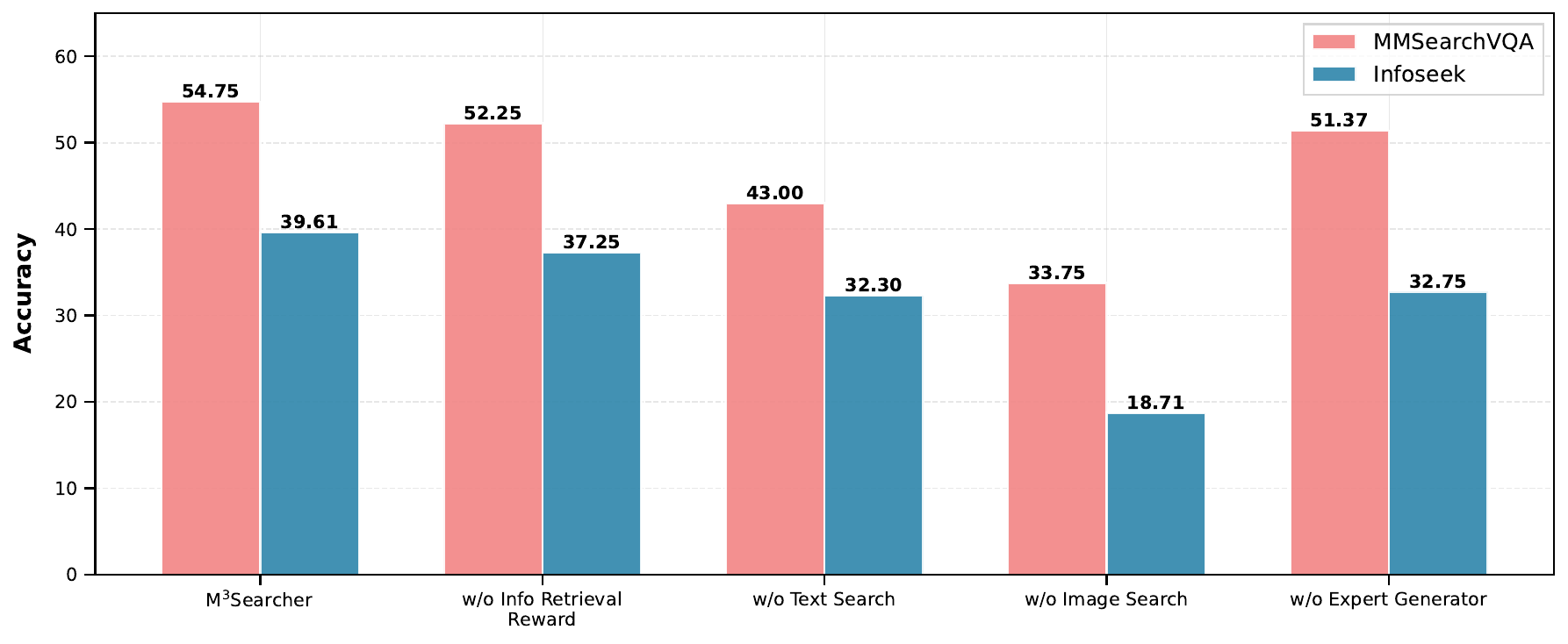}
    \caption{Ablation study.}
    \label{ablation}
\end{figure}

\section{Main Results}
The overall performance of M$^3$Searcher across multiple benchmarks is summarized in Table \ref{main}. Several insights can be drawn from these results:
(1) \textbf{Baseline methodologies exhibit unstable variability in performance across benchmarks.} In most cases, either fixed RAG pipelines or prompt-engineered agents attain the strongest results. This suggests that explicit, hand-crafted prompt engineering provides a competitive advantage that decoupled, untrained agents fail to surpass.
Agents with specialized training display inconsistent performance and unstable generalization: for example, MMSearch-R1 performs competitively on Infoseek, but its performance drops sharply on out-of-distribution tasks.
(2) \textbf{M$^3$Searcher demonstrates robust and strong performance across various generalization and transfer settings.} It provides high-quality, correctly excavated evidence and both multimodal and purely textual backbones can reliably synthesize accurate answers, lifting the modality constraints. Notably, DeepSeek-R1 answer generator emerges as the top performer, underscoring the critical role of the inherent reasoning capability of the backbone model in the overall MRAG system effectiveness.
M$^3$Searcher also maintains stable performance under search-engine transfer, exhibiting no degradation when switching search tools. This robustness highlights its high robustness to variations in the underlying information source, and further indicates that a self-built textual search engine is fully sufficient for on-policy RL training — particularly important given the prohibitive cost of commercial search engines.

\section{Analysis}
% \paragraph{Retrieval-oriented reward and web search tools boosts performance}

\paragraph{Ablation study.}
We evaluate the contribution of each core component in M$^3$Searcher by removing the Information Retrieval Reward, the image search tool, the text search tool and the answer generator tool. As shown in Figure \ref{ablation}, each component provides a measurable performance gain, underscoring their collective importance to the overall system effectiveness.

\begin{figure}[h]
    \centering
    \includegraphics[width=\linewidth]{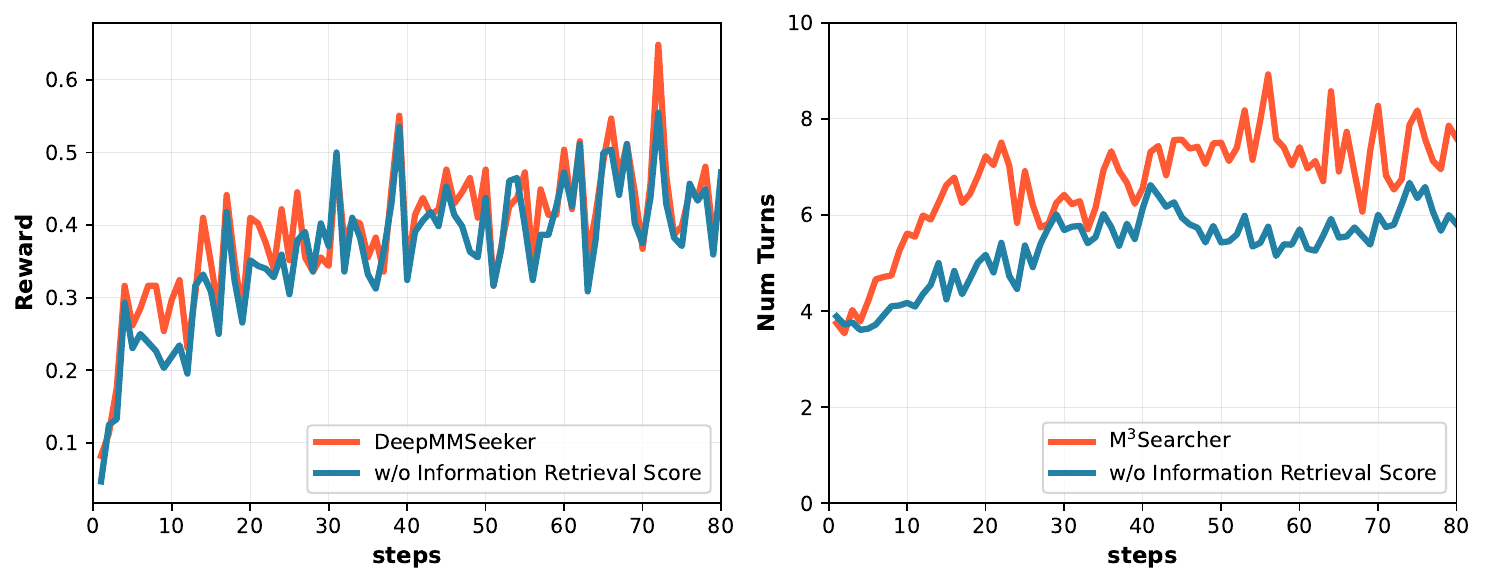}
    \caption{Training dynamics of reward and rollout turn counts with and without the information-retrieval reward.}
    \label{dynamic}
\end{figure}

\begin{figure}[h]
    \centering
    \includegraphics[width=\linewidth]{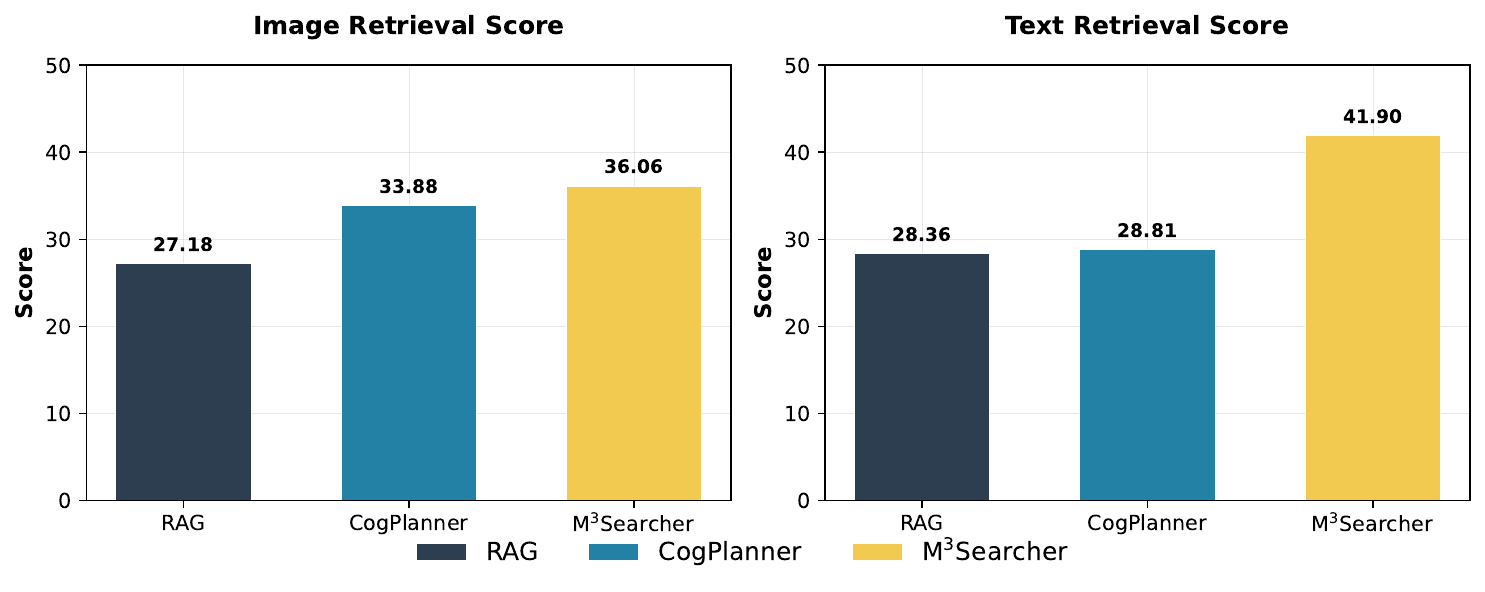}
    \caption{Text retrieval score and image retrieval score of M$^3$Searcher compared with CogPlanner and RAG pipeline baseliness.}
    \label{retrieval}
\end{figure}

\paragraph{Retrieval-oriented rewards enhance the breadth and completeness of information seeking.}
To rigorously evaluate the impact of retrieval-oriented reward design, we analyze the training dynamics presented in Figure \ref{dynamic}. The results indicate that incorporating an information retrieval reward leads to consistently higher reward signals. Consequently, M$^3$Searcher engages in a greater number of information-seeking turns. This enables a broader coverage of relevant information and yielding final evidence that is both more complete and reliable, as demonstrated in Figure \ref{retrieval}.

\begin{figure}[h]
    \centering
    \includegraphics[width=\linewidth]{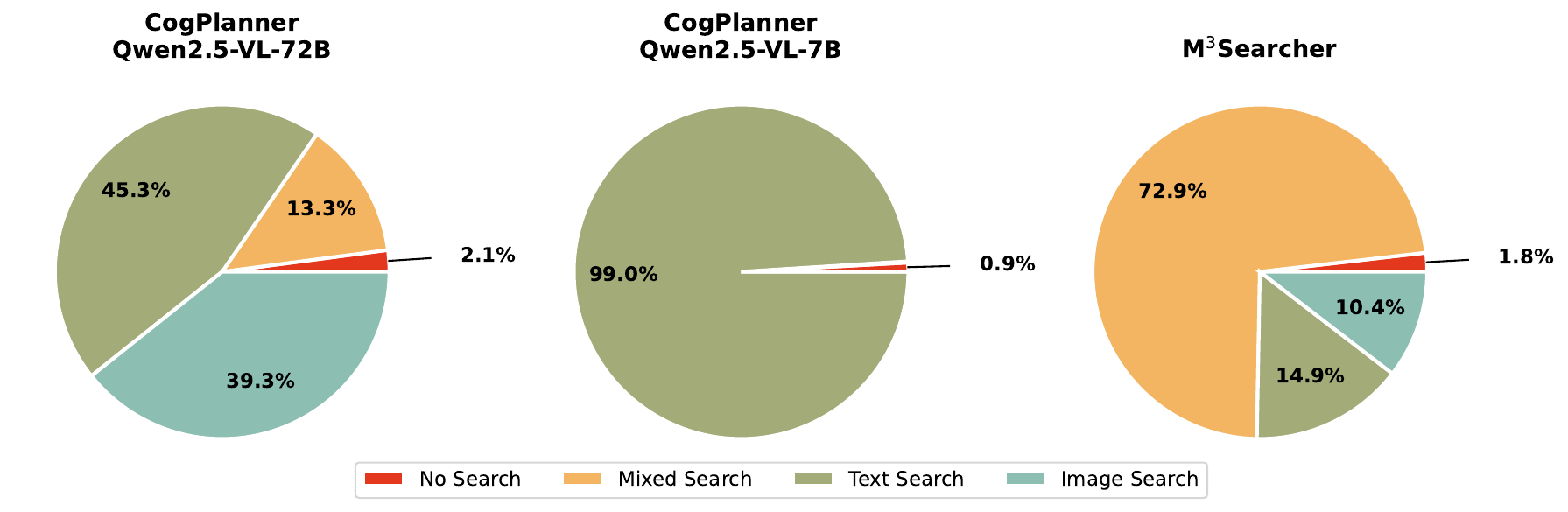}
    \caption{Tool usage statistics on MMSearchVQA.}
    \label{ratio}
\end{figure}

\paragraph{RL enhances heterogeneous tool coordination and improves the model’s ability to leverage the image-search tool} 
We analyze tool usage patterns on the MMSearchVQA benchmark in Figure \ref{ratio}. 
The results reveal a pronounced bias in the pretrained Qwen2.5-VL-7B model as it overwhelmingly favors text search tool while almost never invoking the image search tool. 
% This indicates a fundamental deficiency in its ability to recognize when external visual retrieval is required. 
After RL this imbalance is substantially mitigated. It invokes a more diverse mixture of search actions, with a notably increased reliance on the image-search tool, indicating that RL helps the model internalize when visual external information is necessary for successful reasoning.

% \begin{table*}[t]
%   \centering
%   \caption{The resource consumption statistics.}
%   % \setlength{\tabcolsep}{10pt}
%   \begin{tabular}{lcccc}
%     \toprule[1.3pt]
%     \textbf{Backbone} & \textbf{Method} & \textbf{Token Consumption} &  \multicolumn{2}{c}{\textbf{Search Frequency}} \\ \cmidrule{4-5}
%     & & & \textbf{Image Search} & \textbf{Text Search} \\
%     \midrule 
%     \multirow{2}{*}{Qwen2.5-VL-7B} &   Direct   &   \\
%      &  RAG   &   &   \\
%      &  CogPlanner   &   &   \\
%      &  OmniSearch   &   &   \\
%      \midrule
%     \multicolumn{2}{c}{MMSearch-R1-7B} & \\
%     \midrule
%     \multicolumn{2}{l}{\textbf{MMPlanner-7B} \& Qwen3-32B Generator} &  \\
%     \bottomrule[1.3pt]
%   \end{tabular}
% \end{table*}

\begin{figure}[h]
    \centering
    \includegraphics[width=\linewidth]{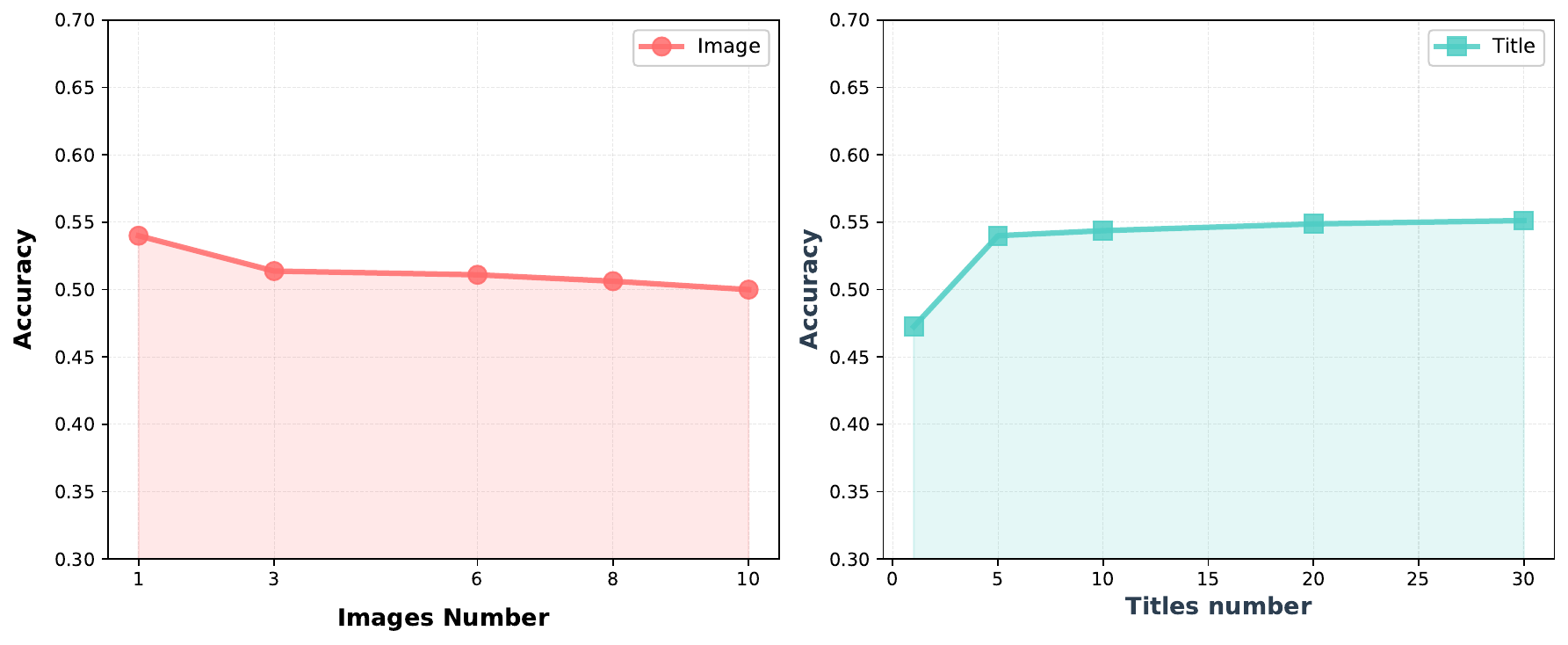}
    \caption{Performance with varying numbers of returned relevant images and associated webpage titles in the image search tool.}
    \label{image_search}
\end{figure}

\paragraph{The Design of Image Search Tools as a Core Factor in Agentic MRAG Performance} 
Our empirical observations indicate that the design of image search tools constitutes a critical determinant of performance. Specifically, we analyzed the influence of the returned relevant images and their associated webpage titles. As depicted in Figure \ref{image_search}, 
% of two distinct information channels returned by the image search tool:
% we conducted an independent analysis varying the quantity of returned relevant images (ranging from 0 to 10) and the quantity of returned webpage titles (ranging from 0 to 50).
% The results reveal a clear divergence in the effect of these information modalities. 
 % and unstable training dynamics
increasing the number of returned images lead to performance degradation, potentially attributed to the redundancy or near-similarity of multiple image inputs, which may introduce noise or confusion into the model's visual feature extraction process. 
% The information seeker does not require a large corpus of visual inputs to infer the context of the search results.
Conversely, increasing the volume of textual returned information (webpage titles) demonstrates a positive correlation with performance since the webpage titles provide crucial context for interpreting the visual query. 
Based on this quantitative analysis, we adopt a design choice for the image search tool that returns the top-1 image along with the top-30 associated webpage titles, which balances informative context with minimal visual redundancy.
% , thereby optimizing the agentic MRAG’s performance.
% Consequently, greater textual input strengthens M$^3$Searcher's capacity for valid information extraction and contextual analysis.
% thereby enhancing M$^3$Searcher’s capacity for valid information extraction and reasoning. Conversely, an excessive number of visually similar images introduces redundancy and noise, which confounds the model and offers limited additional insight regarding the visual query.

% \begin{table*}[t]
%   \centering
%   \caption{The overall performance of MMSearcher on non-linear web search benchmarks.}
%   % \setlength{\tabcolsep}{10pt}
%   \begin{tabular}{lc|cc}
%     \toprule[1.3pt]
%     \textbf{Backbone} & \textbf{Method} & \textbf{?} &  \textbf{MM-BrowseComp} \\
%     % &  Google Search \\ 
%     % \midrule
%     % \multicolumn{3}{l}{\textit{Non Training Agents}} \\
%     \midrule 
%     \multirow{4}{*}{Qwen2.5-VL-72B} &  Direct   &   \\
%      &  RAG   &   &   \\
%      &  CogPlanner   &   &   \\
%      &  OmniSearch   &   &   \\
%     \arrayrulecolor{gray!30}  % 30% gray
%     \midrule
%     \arrayrulecolor{black}
%     \multirow{2}{*}{Qwen2.5-VL-7B} &   Direct   &   \\
%      &  RAG   &   &   \\
%      &  CogPlanner   &   &   \\
%      &  OmniSearch   &   &   \\
%      \midrule
%     \multicolumn{2}{c|}{MMSearch-R1-7B} & \\
%     \midrule
%     \multicolumn{2}{l|}{\textbf{MMPlanner-7B} \& Qwen3-32B Generator} &  \\
%     \bottomrule[1.3pt]
%   \end{tabular}
% \end{table*}

\section{Related Work}
With the introduction of DeepResearch by leading AI organizations, including OpenAI \cite{openai_deep_research_2024}, Google \cite{google_gemini_deep_research_2024}, and Perplexity \cite{perplexity_deepresearch}, these systems have demonstrated strong potential in solving complex multi-step reasoning tasks.
% that require integrating external knowledge. 
% This rapid progress underscores the growing need to enhance LLMs with the ability to efficiently plan, search, and synthesize information from dynamic real-world web environments.
Recent advances highlight reinforcement learning (RL) as a promising paradigm and OpenAI’s technical report explicitly demonstrates the effectiveness of employing RL to strengthen the multi-step decision-making and retrieval abilities \cite{jaech2024openai}. 
Notable works such as Search-R1 \cite{jin2025search,song2025r1} mark an early milestone by incorporating web-search tool interaction into textual question-answering scenarios, achieving substantial performance gains. Following this, the Web Agents series developed by the Qwen team \cite{li2025websailor,wu2025webdancer} further optimizes information-seeking behaviors in complex, non-linear reasoning tasks.
% In parallel, the open-source community has begun exploring end-to-end RL training strategies tailored for agentic RAG systems.
% Notably, works such as Search-R1 \cite{jin2025search,song2025r1} mark an early milestone by incorporating web-search tool interaction into textual question-answering scenarios, achieving substantial performance gains. Following this, the Web Agents series developed by the Qwen team \cite{li2025websailor,wu2025webdancer} further optimizes information-seeking behaviors in complex, non-linear reasoning tasks.
However, despite these advances, rare attention has been given to the optimization of MRAG systems, where reasoning must integrate and synthesize heterogeneous modalities. 
Exsiting work \cite{geng2025webwatcher,wu2025mmsearch,narayan2025deepmmsearch} employ a end-to-end RL paradigm for VQA tasks, which inadvertently restrict the MRAG backbone to relatively small models (e.g., Qwen2.5-VL-7B). This constraint imposes a substantial performance ceiling, limiting the practical effectiveness of these systems in real-world deployments.
% constraining the MRAG system backbone to a limited small scale (e.g. Qwen2.5-VL-7B), largely degrading the overall performance for real application.
% WebWatcher \cite{geng2025webwatcher} and MMSearch-R1 \cite{wu2025mmsearch} 

\section{Conclusion}
We present M$^3$Searcher, a lightweight and trainable multimodal information seeker that decouples retrieval from answer generation in MRAG systems. By focusing on adaptive, reasoning-intensive search over heterogeneous sources, M$^3$Searcher preserves the reasoning capacity of downstream generators while efficiently aggregating contextually relevant evidence. Experiments on MMSearchVQA and real-world benchmarks demonstrate strong performance and robustness across different search engines and generators.

\section*{Limitations}
We discuss several limitations of M$^3$Searcher as follows.
First, although M$^3$Searcher adopts a modular architecture, its effectiveness is inherently constrained by the scale and diversity of the available tool set. Extending the agent to operate over a broader and more heterogeneous collection of real-world tools would substantially enlarge the action space and increase planning complexity.
Second, while MMSearchVQA facilitates retrieval-centric multimodal training, the constructed queries are predominantly characterized by relatively long reasoning trajectories compared to those in existing training corpora. More complex scenarios that require substantially deeper multi-step search and decision-making processes remain underexplored. Extending the dataset construction pipeline therefore represents an important direction for future research.

\bibliography{custom}

@String{Computer = "{IEEE} Computer" }

@article{yu2025unveiling,
  title={Unveiling the potential of multimodal retrieval augmented generation with planning},
  author={Yu, Xiaohan and Yang, Zhihan and Chen, Chong},
  journal={arXiv preprint arXiv:2501.15470},
  year={2025}
}

@article{wu2025webdancer,
  title={WebDancer: Towards Autonomous Information Seeking Agency},
  author={Wu, Jialong and Li, Baixuan and Fang, Runnan and Yin, Wenbiao and Zhang, Liwen and Tao, Zhengwei and Zhang, Dingchu and Xi, Zekun and Fu, Gang and Jiang, Yong and others},
  journal={arXiv preprint arXiv:2505.22648},
  year={2025}
}

@article{wei2025browsecomp,
  title={Browsecomp: A simple yet challenging benchmark for browsing agents},
  author={Wei, Jason and Sun, Zhiqing and Papay, Spencer and McKinney, Scott and Han, Jeffrey and Fulford, Isa and Chung, Hyung Won and Passos, Alex Tachard and Fedus, William and Glaese, Amelia},
  journal={arXiv preprint arXiv:2504.12516},
  year={2025}
}

@article{wu2025mmsearch,
  title={MMSearch-R1: Incentivizing LMMs to Search},
  author={Wu, Jinming and Deng, Zihao and Li, Wei and Liu, Yiding and You, Bo and Li, Bo and Ma, Zejun and Liu, Ziwei},
  journal={arXiv preprint arXiv:2506.20670},
  year={2025}
}

@article{grpo,
  title={Deepseekmath: Pushing the limits of mathematical reasoning in open language models},
  author={Shao, Zhihong and Wang, Peiyi and Zhu, Qihao and Xu, Runxin and Song, Junxiao and Bi, Xiao and Zhang, Haowei and Zhang, Mingchuan and Li, YK and Wu, Yang and others},
  journal={arXiv preprint arXiv:2402.03300},
  year={2024}
}

@article{hu2024mrag,
  title={MRAG-Bench: Vision-Centric Evaluation for Retrieval-Augmented Multimodal Models},
  author={Hu, Wenbo and Gu, Jia-Chen and Dou, Zi-Yi and Fayyaz, Mohsen and Lu, Pan and Chang, Kai-Wei and Peng, Nanyun},
  journal={arXiv preprint arXiv:2410.08182},
  year={2024}
}

@article{jiang2024mmsearch,
  title={Mmsearch: Benchmarking the potential of large models as multi-modal search engines},
  author={Jiang, Dongzhi and Zhang, Renrui and Guo, Ziyu and Wu, Yanmin and Lei, Jiayi and Qiu, Pengshuo and Lu, Pan and Chen, Zehui and Song, Guanglu and Gao, Peng and others},
  journal={arXiv preprint arXiv:2409.12959},
  year={2024}
}

@article{infoseek,
  title={Can pre-trained vision and language models answer visual information-seeking questions?},
  author={Chen, Yang and Hu, Hexiang and Luan, Yi and Sun, Haitian and Changpinyo, Soravit and Ritter, Alan and Chang, Ming-Wei},
  journal={arXiv preprint arXiv:2302.11713},
  year={2023}
}

@article{li2024benchmarkingmultimodalretrievalaugmented,
      title={Benchmarking Multimodal Retrieval Augmented Generation with Dynamic VQA Dataset and Self-adaptive Planning Agent}, 
      author={Yangning Li and Yinghui Li and Xinyu Wang and Yong Jiang and Zhen Zhang and Xinran Zheng and Hui Wang and Hai-Tao Zheng and Pengjun Xie and Philip S. Yu and Fei Huang and Jingren Zhou},
      year={2024},
      eprint={2411.02937},
      archivePrefix={arXiv},
      primaryClass={cs.CL},
      url={https://arxiv.org/abs/2411.02937}, 
}

@article{jin2025search,
  title={Search-r1: Training llms to reason and leverage search engines with reinforcement learning},
  author={Jin, Bowen and Zeng, Hansi and Yue, Zhenrui and Yoon, Jinsung and Arik, Sercan and Wang, Dong and Zamani, Hamed and Han, Jiawei},
  journal={arXiv preprint arXiv:2503.09516},
  year={2025}
}

@article{jaech2024openai,
  title={Openai o1 system card},
  author={Jaech, Aaron and Kalai, Adam and Lerer, Adam and Richardson, Adam and El-Kishky, Ahmed and Low, Aiden and Helyar, Alec and Madry, Aleksander and Beutel, Alex and Carney, Alex and others},
  journal={arXiv preprint arXiv:2412.16720},
  year={2024}
}

@article{song2025r1,
  title={R1-searcher: Incentivizing the search capability in llms via reinforcement learning},
  author={Song, Huatong and Jiang, Jinhao and Min, Yingqian and Chen, Jie and Chen, Zhipeng and Zhao, Wayne Xin and Fang, Lei and Wen, Ji-Rong},
  journal={arXiv preprint arXiv:2503.05592},
  year={2025}
}

@article{li2025websailor,
  title={WebSailor: Navigating Super-human Reasoning for Web Agent},
  author={Li, Kuan and Zhang, Zhongwang and Yin, Huifeng and Zhang, Liwen and Ou, Litu and Wu, Jialong and Yin, Wenbiao and Li, Baixuan and Tao, Zhengwei and Wang, Xinyu and others},
  journal={arXiv preprint arXiv:2507.02592},
  year={2025}
}

@article{geng2025webwatcher,
  title={WebWatcher: Breaking New Frontiers of Vision-Language Deep Research Agent},
  author={Geng, Xinyu and Xia, Peng and Zhang, Zhen and Wang, Xinyu and Wang, Qiuchen and Ding, Ruixue and Wang, Chenxi and Wu, Jialong and Zhao, Yida and Li, Kuan and others},
  journal={arXiv preprint arXiv:2508.05748},
  year={2025}
}

@article{fu2025livevqa,
  title={LiveVQA: Live Visual Knowledge Seeking},
  author={Fu, Mingyang and Peng, Yuyang and Liu, Benlin and Wan, Yao and Chen, Dongping},
  journal={arXiv preprint arXiv:2504.05288},
  year={2025}
}

@article{cheng2025simplevqa,
  title={Simplevqa: Multimodal factuality evaluation for multimodal large language models},
  author={Cheng, Xianfu and Zhang, Wei and Zhang, Shiwei and Yang, Jian and Guan, Xiangyuan and Wu, Xianjie and Li, Xiang and Zhang, Ge and Liu, Jiaheng and Mai, Yuying and others},
  journal={arXiv preprint arXiv:2502.13059},
  year={2025}
}

@article{bai2025qwen25,
  title={Qwen2. 5-vl technical report},
  author={Bai, Shuai and Chen, Keqin and Liu, Xuejing and Wang, Jialin and Ge, Wenbin and Song, Sibo and Dang, Kai and Wang, Peng and Wang, Shijie and Tang, Jun and others},
  journal={arXiv preprint arXiv:2502.13923},
  year={2025}
}

@article{guo2025deepseek,
  title={Deepseek-r1: Incentivizing reasoning capability in llms via reinforcement learning},
  author={Guo, Daya and Yang, Dejian and Zhang, Haowei and Song, Junxiao and Zhang, Ruoyu and Xu, Runxin and Zhu, Qihao and Ma, Shirong and Wang, Peiyi and Bi, Xiao and others},
  journal={arXiv preprint arXiv:2501.12948},
  year={2025}
}

@article{liu2024deepseek,
  title={Deepseek-v3 technical report},
  author={Liu, Aixin and Feng, Bei and Xue, Bing and Wang, Bingxuan and Wu, Bochao and Lu, Chengda and Zhao, Chenggang and Deng, Chengqi and Zhang, Chenyu and Ruan, Chong and others},
  journal={arXiv preprint arXiv:2412.19437},
  year={2024}
}

@article{yang2025qwen3,
  title={Qwen3 technical report},
  author={Yang, An and Li, Anfeng and Yang, Baosong and Zhang, Beichen and Hui, Binyuan and Zheng, Bo and Yu, Bowen and Gao, Chang and Huang, Chengen and Lv, Chenxu and others},
  journal={arXiv preprint arXiv:2505.09388},
  year={2025}
}

@inproceedings{tran2025reasonvqa,
  title={ReasonVQA: A Multi-hop Reasoning Benchmark with Structural Knowledge for Visual Question Answering},
  author={Tran, Duong T and Tran, Trung-Kien and Hauswirth, Manfred and Le Phuoc, Danh},
  booktitle={Proceedings of the IEEE/CVF International Conference on Computer Vision},
  pages={18793--18803},
  year={2025}
}

@article{li2025mm,
  title={Mm-browsecomp: A comprehensive benchmark for multimodal browsing agents},
  author={Li, Shilong and Bu, Xingyuan and Wang, Wenjie and Liu, Jiaheng and Dong, Jun and He, Haoyang and Lu, Hao and Zhang, Haozhe and Jing, Chenchen and Li, Zhen and others},
  journal={arXiv preprint arXiv:2508.13186},
  year={2025}
}

@article{narayan2025deepmmsearch,
  title={DeepMMSearch-R1: Empowering Multimodal LLMs in Multimodal Web Search},
  author={Narayan, Kartik and Xu, Yang and Cao, Tian and Nerella, Kavya and Patel, Vishal M and Shiee, Navid and Grasch, Peter and Jia, Chao and Yang, Yinfei and Gan, Zhe},
  journal={arXiv preprint arXiv:2510.12801},
  year={2025}
}

@article{shao2025reasonir,
  title={ReasonIR: Training Retrievers for Reasoning Tasks},
  author={Shao, Rulin and Qiao, Rui and Kishore, Varsha and Muennighoff, Niklas and Lin, Xi Victoria and Rus, Daniela and Low, Bryan Kian Hsiang and Min, Sewon and Yih, Wen-tau and Koh, Pang Wei and others},
  journal={arXiv preprint arXiv:2504.20595},
  year={2025}
}

@article{wang2022text,
  title={Text embeddings by weakly-supervised contrastive pre-training},
  author={Wang, Liang and Yang, Nan and Huang, Xiaolong and Jiao, Binxing and Yang, Linjun and Jiang, Daxin and Majumder, Rangan and Wei, Furu},
  journal={arXiv preprint arXiv:2212.03533},
  year={2022}
}

@article{jiang2025s3,
  title={s3: You Don't Need That Much Data to Train a Search Agent via RL},
  author={Jiang, Pengcheng and Xu, Xueqiang and Lin, Jiacheng and Xiao, Jinfeng and Wang, Zifeng and Sun, Jimeng and Han, Jiawei},
  journal={arXiv preprint arXiv:2505.14146},
  year={2025}
}

@article{kalajdzievski2024scaling,
  title={Scaling laws for forgetting when fine-tuning large language models},
  author={Kalajdzievski, Damjan},
  journal={arXiv preprint arXiv:2401.05605},
  year={2024}
}

@article{li2024revisiting,
  title={Revisiting catastrophic forgetting in large language model tuning},
  author={Li, Hongyu and Ding, Liang and Fang, Meng and Tao, Dacheng},
  journal={arXiv preprint arXiv:2406.04836},
  year={2024}
}

@misc{zheng2025deepresearcherscalingdeepresearch,
      title={DeepResearcher: Scaling Deep Research via Reinforcement Learning in Real-world Environments}, 
      author={Yuxiang Zheng and Dayuan Fu and Xiangkun Hu and Xiaojie Cai and Lyumanshan Ye and Pengrui Lu and Pengfei Liu},
      year={2025},
      eprint={2504.03160},
      archivePrefix={arXiv},
      primaryClass={cs.AI},
      url={https://arxiv.org/abs/2504.03160}, 
}

@article{sheng2024hybridflow,
  title   = {HybridFlow: A Flexible and Efficient RLHF Framework},
  author  = {Guangming Sheng and Chi Zhang and Zilingfeng Ye and Xibin Wu and Wang Zhang and Ru Zhang and Yanghua Peng and Haibin Lin and Chuan Wu},
  year    = {2024},
  journal = {arXiv preprint arXiv: 2409.19256}
}

@article{openai_deep_research_2024,
  title={OpenAI Deep Research System Card},
  author={OpenAI},
  journal={OpenAI Blog},
  year={2025},
}

@article{google_gemini_deep_research_2024,
  title={Gemini Deep Research System Card},
  author={Google},
  institution={Google},
  year={2025},
}

@article{perplexity_deepresearch,
  title={Perplexity Deep Research System Card},
  author={Perplexity},
  institution={Perplexity},
  year={2025},
}

\appendix

\section{Implementation Details}
\label{implementation}
\subsection{RAG}
For the RAG baseline, we adopt the prompt design and processing pipeline proposed in \cite{wu2025mmsearch}. Specifically, we first perform image retrieval using the Serper API, followed by query refinement based on the image search context. The refined query is then used to conduct a subsequent text search, and the final response is generated using the combined information.
It is worth noting that our implementation differs slightly from \cite{wu2025mmsearch} in that we do not utilize the Jina API to fetch the full webpage content.

\subsection{OmniSearch}
For the OmniSearch baseline, we leverage the publicly available implementation\footnote{\url{https://github.com/Alibaba-NLP/OmniSearch}} and adapt the search tool interface to match our experimental setup. Apart from this minor adjustment, we retain the original workflow and prompt design to ensure a fair comparison.

\subsection{CogPlanner}
For CogPlanner, we develop a multi-agent planning framework built upon the \texttt{llama-index} library\footnote{\url{https://github.com/run-llama/llama_index}}. This implementation integrates dynamic query reformulation and retrieval strategy selection to facilitate efficient multimodal information synthesis.

\subsection{MMSearch-R1}
% We utilize two version of implementation. First, we utilize the publicly available codes\footnote{\url{https://github.com/EvolvingLMMs-Lab/multimodal-search-r1}} and replace the image search and text search with our own built tools. However, the tool call quickly drop to near zero and is rarely invoked. Thus, we also pull the released model checkpoint\footnote{\url{https://huggingface.co/lmms-lab/MMSearch-R1-7B/tree/main}} for evaluation.

We adopt two implementation strategies. First, we build upon the publicly available codes\footnote{\url{https://github.com/EvolvingLMMs-Lab/multimodal-search-r1}}
and substitute the original image and text search components with our custom-built tools. However, under this setting, the model’s tool invocation frequency rapidly diminishes to nearly zero. 
To ensure a fair and stable evaluation, we additionally employ the released model checkpoint\footnote{\url{https://huggingface.co/lmms-lab/MMSearch-R1-7B}} for our experiments.

\section{Prompts}
\label{rollout}

\begin{tcolorbox}[
  title={Multimodal RAG Rollout}, 
  width=0.5\textwidth,
  colback=blue!5,
  colframe=darkblue,
  coltitle=white
]
\label{juding}
You are an expert in information seeking and reasoning. You will be given a question with a image. You need to collect information for the question step by step. \newline
Follow these instructions carefully: \newline
1. If you need external knowledge, call search tools. \newline
2. Enclose your entire reasoning process within <think> ... </think> tags.\newline
3. If you find no further external knowledge needed, stop the search process and call the answer model tool.
\end{tcolorbox}

\begin{tcolorbox}[title = {Answer Reward}, width=0.5\textwidth]
\label{juding}
You are an expert evaluator. You will be given: \newline
- A question \newline
- Several correct (golden) answer candidates\newline
- My provided answer\newline

Your task:\newline
Strictly judge whether my answer is correct compared to the golden answers.\newline

Judgement rules:\newline
1. The meaning of my answer **must match** one of the golden answer candidates.\newline
2. Reject or fail to answer is wrong answer.\newline

Output format:\newline
<reason> The reason of judgement. <Judgement> Yes or No.\newline

Question: \{question\}\newline
Golden Answer: \{cand\_ans\}\newline
My Answer: \{gen\}

\end{tcolorbox}

\end{document}